\pdfoutput=1

\documentclass[11pt]{article}

\usepackage[]{acl}

\usepackage{times}
\usepackage{latexsym}
\usepackage{booktabs}
\usepackage{xcolor,colortbl}
\usepackage{diagbox}
\usepackage{graphicx}
\usepackage{amsmath}
\usepackage{xfp}
\usepackage{tabulary}
\usepackage{threeparttablex}
\usepackage{multicol}
\usepackage{array}
\usepackage{amsthm,amsmath,amssymb}
\usepackage{tabularx}
\usepackage{mathrsfs}
\usepackage{arydshln}
\usepackage{paralist}

\usepackage[T1]{fontenc}

\usepackage[utf8]{inputenc}

\usepackage{microtype}

\usepackage{color, xcolor}

\newcommand{\darkercell}{\cellcolor{blue!10!white}}
\newcommand{\lightercell}{\cellcolor{blue!5!white}}
\newcommand{\orgone}{\textsuperscript{$\diamondsuit$}}
\newcommand{\orgtwo}{\textsuperscript{$\heartsuit$}}
\newcommand{\orgthree}{\textsuperscript{$\spadesuit$}}

\renewcommand{\thefootnote}{\fnsymbol{footnote}}
\title{Is There a One-Model-Fits-All Approach to Information Extraction? Revisiting Task Definition Biases}

\author{Wenhao Huang\orgone, Qianyu He\orgone, Zhixu Li\orgone, Jiaqing Liang\orgtwo \footnote[2]{}, Yanghua Xiao\orgone\orgthree\footnote[2]{}\\
\orgone Shanghai Key Laboratory of Data Science, School of Computer Science, Fudan University \\
\orgtwo School of Data Science, Fudan University \\
\orgthree Fudan-Aishu Cognitive Intelligence Joint Research Center\\
\texttt{\{whhuang21,qyhe21\}@m.fudan.edu.cn} \\ 
\texttt{\{liangjiaqing,zhixuli,shawyh\}@fudan.edu.cn} \\
}

\begin{document}
\maketitle
\footnotetext[2]{Corresponding authors.}

\renewcommand*{\thefootnote}{\arabic{footnote}}
\begin{abstract}
\textit{Definition bias} is a negative phenomenon that can mislead models. Definition bias in information extraction appears not only across datasets from different domains but also within datasets sharing the same domain. We identify two types of definition bias in IE: bias among information extraction datasets and bias between information extraction datasets and instruction tuning datasets. To systematically investigate definition bias, we conduct three probing experiments to quantitatively analyze it and discover the limitations of unified information extraction and large language models in solving definition bias. To mitigate definition bias in information extraction, we propose a multi-stage framework consisting of definition bias measurement, bias-aware fine-tuning, and task-specific bias mitigation. Experimental results demonstrate the effectiveness of our framework in addressing definition bias.~\footnote{Resources of this paper can be found at \url{https://github.com/EZ-hwh/definition-bias}}
\end{abstract}

\section{Introduction}
Bias in machine learning refers to systematic errors in predictions in the machine learning process, such as annotator bias, measurement bias, etc~\cite{hellström2020bias}. 
In the era of large language models (LLMs), this issues are addressed by filtering low-quality corpora~\cite{kojima2022large} and training with human preferences~\cite{ouyang2022training}. 
However, performance remains subpar in handling information extraction (IE) tasks~\cite{wadhwa2023revisiting}, which we believe is due to definition bias.

\begin{figure}[t]
	\centering
	\includegraphics[width=\linewidth]{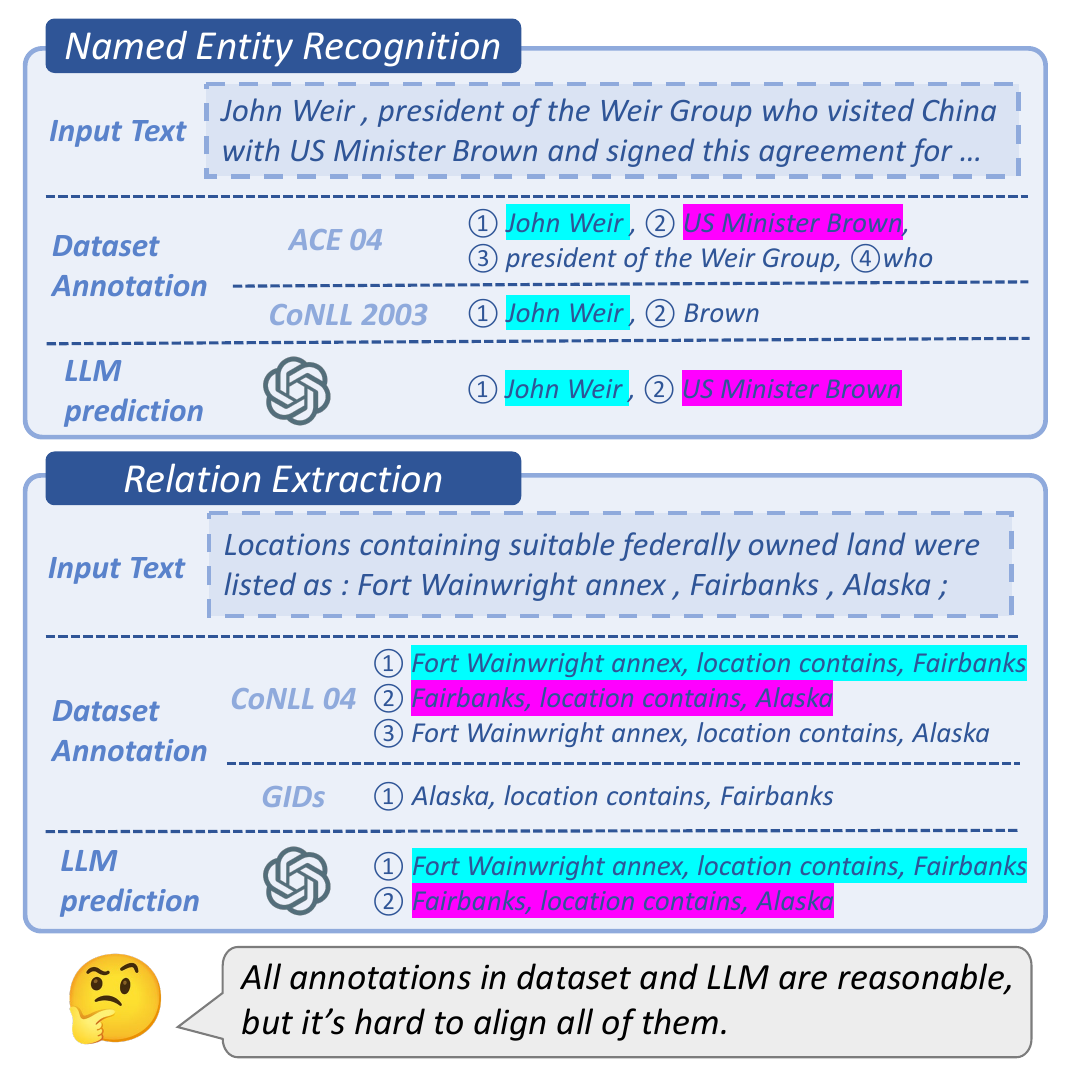}
	\caption{Definition bias among different datasets and LLMs even when they share the same entity type (for NER) or the same relation type (for RE).}
	\label{fig:intro}
\end{figure}

Definition bias in IE refers to the tendency of an information extraction system to favor certain interpretations of data over others, often due to the way concepts, entities, or relationships are defined within the system. 
As the fast development of Unified Information Extraction (UIE)~\cite{lu-etal-2022-unified} and Large Language Models (LLMs)~\cite{openai2022chatgpt, openai2023gpt4, geminiteam2023gemini} in recent years, two novel definition bias emerge, which are:
\textit{Bias among IE datasets} and \textit{Bias between IE and instruction tuning (IFT) datasets}. Regarding \textit{Bias among IE datasets}, it refers to the definition differences between different data sets under the same annotation schema. As illustrated in Figure~\ref{fig:intro}, different datasets have different annotations to the same input for both Named Entity Recognition (NER) and Relation Extraction (RE) tasks.
Regarding \textit{Bias between IE and instruction tuning datasets}, it highlights the mismatch between the information extraction task and the general task. As depicted in Figure~\ref{fig:intro}, although GPT-4~\cite{openai2023gpt4} is capable of extracting entities or relational triples in accordance with the specified task description without providing extra examples, its prediction differ from those in the existing datasets.

\begin{figure*}[!ht]
	\centering
	\includegraphics[width=\linewidth]{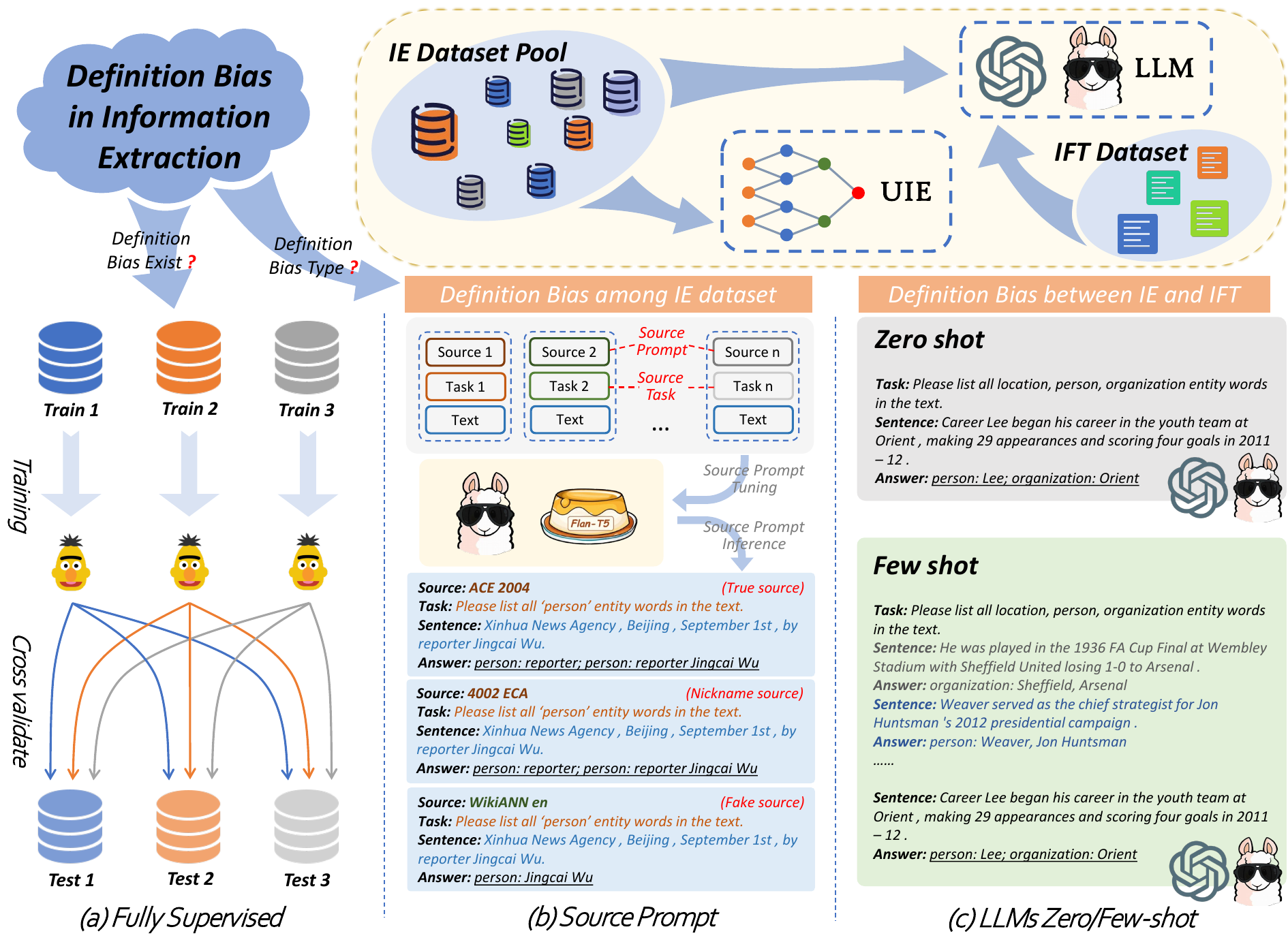}
	\caption{Three settings for the probing tasks on definition bias across datasets, including (a) fully supervised, (b) source prompt and (c) LLMs zero/few-shot.}
	\label{fig:probing_task}
\end{figure*}

To systematically investigate definition bias in IE, we devise a series of probing experiments. 
First, we analyze \textit{whether definition bias exists and how it varies among datasets sharing the same tasks}. 
By conducting cross-validation experiment among various dataset in the NER and RE, we observe a significant decrease in performance, indicating that definition bias negatively impacts the transferability of a fully-supervised model. 
An intuitive way to alleviate definition bias is unified information extraction, which is trained across multiple IE dataset.
Therefore, we analyze \textit{in the unified information extraction setting, does definition bias still exist?}
By introducing source prompt~\cite{li2022learning} that apply true or fake source name for the UIE models, we discover the inconsistency of the UIE for extraction, which indicates that UIE suffers from definition bias among IE datasets.
The other way to mitigate definition bias is LLMs, which is able to understand a wide range of human instructions.
Thereupon, we analyze \textit{Can LLMs address the challenge of definition bias?}
By conducting experiments on few-shot settings on NER and RE task with in-context learning, we find that it's difficult for LLMs without parameter updates to attain satisfactory performance, which indicates that LLMs still suffer from definition bias between IE and instruction fine-tuning datasets.

According to our probing experiments, it is imperative to address definition bias by proposing a universal solution for IE tasks.
However, mitigating definition bias is non-trivial, primarily owing to the following three challenges.
\begin{inparaenum}[(1)]
    \item Enhancing the capacity of LLMs in general information extraction tasks is vital to reduce the definition bias between information extraction datasets and instruction tuning datasets;
    \item Mitigating the definition bias during the tuning of LLMs with different IE datasets;
    \item Learning from new data over time, adapting to new tasks while ensuring the model remains good performance on existing task, is a significant challenge.
\end{inparaenum}

To address these challenges, we propose a framework to alleviate definition bias, which consists of definition bias measurement, bias-aware fine-tuning and task-specific bias mitigation. Using Fleiss's Kappa~\cite{fleiss1971measuring}, we measure the two types of definition bias above. Then we conduct bias-aware fine-tuning with multiple information extraction instructions to enhance the  extraction capabilities with less definition bias. Ultimately, we conduct the task-specific bias mitigation, with low-rank adaptation technique (LoRA)~\cite{hu2021lora} for specific information extraction tasks to further align the LLMs with annotations.

Our paper is organized as follows: In Section~\ref{sec:probing_setting}, we present three probing experiments designed to explore the presence of definition bias and assess the ability of existing frameworks to address this issue in IE. Section~\ref{sec:probing_exp} details the results and analysis of these experiments, concluding that frameworks based on either one-stage processing or parameter-free updates are insufficient to tackle definition bias in IE. Consequently, we propose a novel framework featuring two-stage fine-tuning, specifically developed to mitigate the identified definition bias, as introduced in Section~\ref{sec:framework}. Ultimately, in Section~\ref{sec:exp}, we compare the performance of our framework with state-of-the-art methods in universal information extraction, demonstrating its effectiveness in reducing definition bias.

\section{Definition Bias Probing Experiment}\label{sec:probing_task}\label{sec:probing_setting}
\newtheorem{assumption}{Assumption}
We initially propose an experiment employing cross-validation to investigate the presence of definition bias in the IE tasks. Subsequently, we design two specific detection tasks: source prompt detection and few-shot prompting in LLMs, to examine two categories of definition bias: bias within IE datasets and bias between IE and instruction fine-tuning datasets. These experiments aim to explore the effectiveness of the UIE and LLM frameworks in addressing the definition bias issue.


\subsection{Whether definition bias exists?}
\label{FullySupervisedSettings}
To better illustrate the definition bias among different information extraction tasks, we design a cross-extraction task. 
As shown in Figure~\ref{fig:probing_task}(a), we train multiple fully-supervised models with different datasets on the same task (NER and RE) respectively, and test them on other datasets to evaluate whether definition bias exists.



We first introduce two BERT-based extraction frameworks to handle the NER and RE task, respectively.

\paragraph{Named Entity Recognition} We adopt GlobalPointer~\cite{su2022global}, an efficient span-based approach that models the beginning and end positions to predict entities using a two-dimensional scoring matrix. By incorporating extended softmax and cross-entropy, GlobalPointer is better equipped to learn from scenarios involving class imbalance.

\paragraph{Relation Extraction} We adopt \textsc{ReRe}~\cite{xie-etal-2021-revisiting} as the basic model for relation extraction. \textsc{ReRe} is a pipeline approach that first performs sentence-level relation detection, followed by subject/object extraction. Specifically, the \textsc{ReRe} model treats the former as a multi-class classification task and the latter as a span detection task.

During the cross-validation process, we encountered label type biases across different datasets. For instance, the ACE 2004 dataset requires the extraction of the \texttt{weapon} entity, which is not a requirement in the CoNLL 2003 dataset. Consequently, we focus exclusively on the types of labels (such as entity types in NER and relationship types in RE) that are annotated in both the training and testing datasets. An example is the \texttt{person} label, which is common for both ACE 2004 and CoNLL 2003.

To mitigate the impact of text distribution shift on the experimental results, we sample a subset of sentences with similar semantics as a cross-validation set. Specifically, we measure the semantic similarity between two sentences by calculating the cosine similarity of their sentence embeddings.  We define the semantic similarity of the sentence $sent_i$ to the dataset $\mathcal{D}$. Finally, we filter out all sentences that fall below $threshold(\mathcal{D})$.

{
    \fontsize{10pt}{12pt}\selectfont
    \begin{gather}%
        sim(sent_i, \mathcal{D}) = \max_{ref_j\in\mathcal{D}} cosine(V_{sent_i}, V_{ref_j}) \\
        threshold(\mathcal{D}) = \sigma \cdot \frac{1}{|\mathcal{D}|}\sum_{s_i\in\mathcal{D}} sim(s_i, \mathcal{D}\setminus \{s_i\})
    \end{gather}
}

where $V_{S}$ denotes the embedding vector of a sentence $S$ encoded by a sentence model\footnote{We adopt MPNet~\cite{song2020mpnet} as our sentence embedding encoder, which is commonly used for retrieval.}, and $\sigma$ denotes the hyperparameters that adjust the threshold, empirically set to $0.7$.  

\subsection{Can UIE address definition bias?} \label{SourcePromptSettings}
Unified information extraction, which employs a pre-defined structured extraction language to encode different extraction structures, can accurately recognize extraction instructions. Inspired by \citet{li2022learning}, which introduces a novel prompt-based method in a transferable setting for text generation tasks, we adopt source prompt settings for probing. Briefly, in our experiment setting, a source can be denoted as the name of the dataset (e.g., ACE 2004). By presenting UIE with various sources—indicating which dataset the instance is from—we can guide it to yield different extraction results. This approach allows us to assess whether it can maintain consistent results with different source prompts.

As Figure~\ref{fig:probing_task}(b) shows, the probing experiment consists of two parts: \textit{source prompt tuning} and \textit{source prompt inference}. Initially, we undertake a source prompt tuning process to enhance the UIE model's ability to recognize different sources. Subsequently, we examine the definition bias within the UIE model by introducing various sources.

\paragraph{Source Prompt Tuning}
The source prompt process can be regarded as a general multi-task learning framework. First, we define a set of source information extraction tasks $\mathcal{S}=\{\mathcal{S}_1,...,\mathcal{S}_n\}$, where the $k$-th task $\mathcal{S}_k=\{(x_i^k, y_i^k)\}_{i=1}^{N_k}$ contains $N_k$ tuples of the input text $x_i^k\in\mathcal{X}_k$ and its corresponding output text $y_i^k\in\mathcal{Y}_k$. For a target information extraction task $\mathcal{T}$, the goal of multi-task learning is to leverage previously learned task-specific knowledge of the source tasks $\mathcal{S}$ to improve prediction of the extraction result. Unlike the traditional multitask fine-tuning scenario, in source prompt tuning, we learn an independent \textit{source prompt} $p_k$ for each source information extraction task $\mathcal{S}_k$ in source prompt tuning, where $x_i^k$ consists of extraction task source name $s_k$, information extraction task description $t_k$, and the sentence $sent_i^k$.
For example, a single instance \textit{"Here's a dataset from ACE 2004, please list all 'person' entity words in the text. Input sentence: Xinhua News Agency, Beijing, September 1st, by reporter Jingcai Wu."} contains the components that are described above.

To clearly demonstrate that UIE with instruction tuning can implicitly learn the definitions of dataset through source prompt, we assign a nickname $p_k'$ for every dataset and randomly replace $p_k$ with $p_k'$. For simplicity, we merely reverse the order of the original dataset names, thereby generating a non-natural language nicknames. For example, the dataset name \textit{"ACE 2004"} is replaced with \textit{"4002 ECA"}. This procedure is designed to eliminate the influence caused by the differences in learning various source names in the UIE and to ensure that the discrepancies in results between true and fake settings are solely due to dataset definition bias.

Specifically, we adopt Llama-v2-13B~\cite{touvron2023llama} and FlanT5-11B~\cite{chung2022scaling} as our backbone models in source prompt tuning settings because of their powerful instruction understanding and instruction-following capabilities.
Based on multiple datasets in NER and RE, we add an additional \textit{source prompt} to every extraction instance to indicate the dataset to which it belongs. Further details on source prompt tuning are described in the Appendix~\ref{apd:source_prompt}.

\paragraph{Source Prompt Inference}
In reference, we provide different source prompts with the same extraction instance to our UIE models that have been fine-tuned on the dataset with source prompts. To probe the definition bias in universal generative information extraction, UIE predicts the extraction result with \textit{\textbf{True source}} (the extraction case with the original source name), \textit{\textbf{Nickname source}} (a nickname of the original source name) and \textit{\textbf{Fake source}} (the extraction case with a fake source name). With different source name, UIE generates different extraction results following different definitions learned from source prompt tuning.

\subsection{Can LLMs address definition bias?}
Large language models exhibit remarkable instruction understanding capabilities, which help them achieve extraordinary performance on various tasks. However, due to the definition bias between IE datasets and IFT datasets, there is a significant performance gap in LLMs when it comes to the information extraction task~\cite{wadhwa2023revisiting}. In-context learning, where LLMs make predictions based solely on contexts augmented with a few examples, is a training-free learning framework that enables models to adapt to new tasks~\cite{dong2023survey}. It is considered a solution to address the definition bias between IE datasets and instruction tuning datasets.

As shown in Figure~\ref{fig:probing_task}(c), we conduct the probing experiment with multiple LLMs in both zero-shot and few-shot settings.

Specifically, we utilize open-source LLMs such as Llama-v2-chat-70B~\cite{touvron2023llama}, and close-source LLMs GPT-3.5-Turbo~\cite{openai2022chatgpt}, GPT-4~\cite{openai2023gpt4} as our backbone models. 
In zero-shot settings, we prompt LLMs with a task description, which probes the definition bias between IE and IFT datasets. Meanwhile, in few-shot settings, we prompt LLMs with a task description and an additional four cases randomly sampled from the corresponding training set to examine whether in-context learning can address the definition bias. For a fair comparison, we sample 200 cases from each dataset and test them in both zero-shot and few-shot settings, respectively.

\section{Empirical Study of Definition Bias}\label{sec:probing_exp}

\renewcommand{\arraystretch}{1.6}
\begin{table}[t]
    \centering
    \resizebox{\columnwidth}{!}{
    \begin{threeparttable}
    
    \begin{tabular}{lcccccccc}
    \toprule
          & \darkercell \textbf{A04}\tnote{1} & \lightercell \textbf{A05}\tnote{2} & \darkercell \textbf{C03}\tnote{3} & \lightercell \textbf{Ont}\tnote{4} & \darkercell \textbf{Wie}\tnote{5} & \lightercell \textbf{TN7}\tnote{6} & \darkercell \textbf{WiN}\tnote{7} & \lightercell \textbf{PoN}\tnote{8} \\
          \darkercell \textbf{A04}\tnote{1} & \cellcolor{purple!\fpeval{8510/85.10}}85.10 & \cellcolor{purple!\fpeval{8219/86.59}}82.19 & \cellcolor{purple!\fpeval{3577/93.78}}35.77 & \cellcolor{purple!\fpeval{2889/94.04}}28.89 & \cellcolor{purple!\fpeval{4989/87.35}}49.89 & \cellcolor{purple!\fpeval{2806/75.75}}28.06 & 
          \cellcolor{purple!\fpeval{3054/95.29}}30.54 &
          \cellcolor{purple!\fpeval{1764/78.13}}17.64 \\
          
         \lightercell \textbf{A05}\tnote{2} & \cellcolor{purple!\fpeval{8344/85.86}}83.44 & \cellcolor{purple!\fpeval{8445/84.45}}84.45 & \cellcolor{purple!\fpeval{3780/93.44}}37.80 & \cellcolor{purple!\fpeval{2643/93.49}}26.43 & \cellcolor{purple!\fpeval{4653/86.71}}46.53 & \cellcolor{purple!\fpeval{2694/78.70}}26.94 &
         \cellcolor{purple!\fpeval{2909/95.52}}29.09 &
         \cellcolor{purple!\fpeval{1823/76.36}}18.23\\
         
         \darkercell \textbf{C03}\tnote{3} & \cellcolor{purple!\fpeval{2410/83.84}}24.10 & \cellcolor{purple!\fpeval{1657/83.10}}16.57 & \cellcolor{purple!\fpeval{9219/92.19}}92.19 & \cellcolor{purple!\fpeval{5582/94.50}}55.82 & \cellcolor{purple!\fpeval{5510/87.85}}55.10 & \cellcolor{purple!\fpeval{7826/80.85}}78.26 &
         \cellcolor{purple!\fpeval{9208/96.14}}92.08 &
          \cellcolor{purple!\fpeval{5367/85.40}}53.67 \\
         
         \lightercell \textbf{Ont}\tnote{4} & \cellcolor{purple!\fpeval{3253/85.47}}32.53 & \cellcolor{purple!\fpeval{2120/84.04}}21.20 & \cellcolor{purple!\fpeval{6060/93.65}}60.60 & \cellcolor{purple!\fpeval{8969/89.69}}89.69 & \cellcolor{purple!\fpeval{4976/87.44}}49.76 & \cellcolor{purple!\fpeval{3475/70.33}}34.75 &
         \cellcolor{purple!\fpeval{6123/95.01}}61.23 &
          \cellcolor{purple!\fpeval{3758/76.60}}37.58 \\
         
         \darkercell \textbf{Wie}\tnote{5} & \cellcolor{purple!\fpeval{2309/84.17}}23.09 & \cellcolor{purple!\fpeval{842/84.50}}8.42 & \cellcolor{purple!\fpeval{6710/94.53}}67.10 & \cellcolor{purple!\fpeval{4114/90.55}}41.14 & \cellcolor{purple!\fpeval{8660/86.60}}86.60 & \cellcolor{purple!\fpeval{6199/77.88}}61.99 &
         \cellcolor{purple!\fpeval{7096/95.99}}70.96 &
          \cellcolor{purple!\fpeval{4413/79.36}}44.13 \\
         
         \lightercell \textbf{TN7}\tnote{6} & \cellcolor{purple!\fpeval{2560/90.20}}25.60 & \cellcolor{purple!\fpeval{2107/85.32}}21.07 & \cellcolor{purple!\fpeval{7616/94.83}}76.16 & \cellcolor{purple!\fpeval{5615/94.53}}56.15 & \cellcolor{purple!\fpeval{7395/89.75}}73.95 & \cellcolor{purple!\fpeval{6339/63.39}}63.39 &
         \cellcolor{purple!\fpeval{8270/96.08}}82.70 &
          \cellcolor{purple!\fpeval{5445/77.62}}54.45\\

         \darkercell \textbf{WiN}\tnote{7} &
         \cellcolor{purple!\fpeval{2548/85.71}}25.48 &
         \cellcolor{purple!\fpeval{2061/86.48}}20.61 &
         \cellcolor{purple!\fpeval{8010/93.92}}80.10 &
         \cellcolor{purple!\fpeval{5869/92.16}}58.69 &
         \cellcolor{purple!\fpeval{5733/87.00}}57.33 &
         \cellcolor{purple!\fpeval{6344/77.09}}63.44 &
         \cellcolor{purple!\fpeval{9521/95.21}}95.21 &
          \cellcolor{purple!\fpeval{5196/77.77}}51.96\\

         \lightercell \textbf{PoN}\tnote{8} &
         \cellcolor{purple!\fpeval{1458/85.00}}14.58 &
         \cellcolor{purple!\fpeval{1084/85.99}}10.84 &
         \cellcolor{purple!\fpeval{4436/93.96}}44.36 &
         \cellcolor{purple!\fpeval{3528/91.98}}35.28 &
         \cellcolor{purple!\fpeval{4026/86.60}}40.26 &
         \cellcolor{purple!\fpeval{3265/78.28}}32.65 &
         \cellcolor{purple!\fpeval{6966/95.38}}69.66 &
         \cellcolor{purple!\fpeval{7777/77.77}}77.77 \\
         
         \bottomrule
    \end{tabular}
    
    \begin{tablenotes}
    \setlength{\multicolsep}{0cm}
    \begin{multicols}{4}
        \item[1] ACE 2004 
        \item[2] ACE 2005 
        \item[3] CoNLL 2003
        \item[4] Ontonotes 
        \item[5] WikiANN en 
        \item[6] TweetNER 7
        \item[7] WikiNeural 
        \item[8] PolyglotNER
    \end{multicols}
    \end{tablenotes}
    \end{threeparttable}
    }
    \caption{Definition bias among different NER tasks.}
    \label{tab:ner_fs}
\end{table}

\renewcommand{\arraystretch}{1.55}
\begin{table}[t]
    \resizebox{\columnwidth}{!}{
    \begin{tabular}{lccccc}
        \toprule
          & \darkercell \textbf{CoNLL 04} & \lightercell \textbf{NYT10} & \darkercell \textbf{NYT11} & \lightercell \textbf{GIDs} & \darkercell \textbf{WikiKBP} \\
          \darkercell \textbf{CoNLL 04} & \cellcolor{purple!\fpeval{6112/61.12}}61.12 & \cellcolor{purple!\fpeval{1020/88.25}}10.20 & \cellcolor{purple!\fpeval{1207/69.00}}12.07 & - & \cellcolor{purple!26.98}26.98 \\
          
         \lightercell \textbf{NYT10} & \cellcolor{purple!\fpeval{1436/58.45}}14.36 & \cellcolor{purple!\fpeval{8968/89.68}}89.68 & \cellcolor{purple!\fpeval{5229/53.33}}52.29 & \cellcolor{purple!\fpeval{1433/62.16}}14.33 & \cellcolor{purple!\fpeval{3032/39.13}}30.32 \\
         
         \darkercell \textbf{NYT11} & \cellcolor{purple!\fpeval{878/59.28}}8.78 & \cellcolor{purple!\fpeval{8332/88.66}}83.32 & \cellcolor{purple!\fpeval{5682/56.82}}56.82 & \cellcolor{purple!\fpeval{1070/62.51}}10.70 & \cellcolor{purple!\fpeval{3264/38.13}}32.64 \\
         \lightercell \textbf{GIDs} & - & \cellcolor{purple!\fpeval{777/55.49}}7.77 & \cellcolor{purple!\fpeval{645/25.00}}6.45 & \cellcolor{purple!\fpeval{6512/65.12}}65.12 & \cellcolor{purple!\fpeval{4964/49.64}}55.65 \\
         
         \darkercell \textbf{WikiKBP} & 0.00 & \cellcolor{purple!\fpeval{1505/59.46}}15.05 & \cellcolor{purple!\fpeval{253/67.33}}2.53 & \cellcolor{purple!\fpeval{2649/70.43}}26.49 & \cellcolor{purple!\fpeval{3657/36.57}}36.57 \\
         
         \bottomrule
    \end{tabular}
    }
    \caption{Definition bias among different RE tasks. Cells with (-) indicates that there are no same relation types between the two datasets.}
    \label{tab:re_fs}
\end{table}
\subsection{Whether definition bias exists?}
Following the cross validation setting described in Section~\ref{FullySupervisedSettings}, we conduct experiment separately on NER and RE tasks in general domain. 
Table~\ref{tab:ner_fs},\ref{tab:re_fs} show the validation result in fully-supervised settings. 

Briefly, we define the model trained and tested on the same dataset as the \textit{reference model}. The numbers in the table cells represent the F1 scores when compared to the golden label. Additionally, the depth of color in each cell indicates the relative quality of extraction in comparison to the reference model. In other words, the darker the cell color, the closer the extraction results are to those of the reference model. The rows of the table represent the training dataset, while the columns represent the test dataset.

Intuitively, the deepest red cells are distributed along the diagonal of the entire table, illustrating that definition bias exists among different datasets, even though they share the same types. This is particularly evident in NER tasks, where several datasets focus on common entity types such as \texttt{person}, \texttt{location}, and \texttt{date}. Despite these similarities, definition bias can lead to significant variations in the model's extraction capabilities.

\begin{figure*}[!t]
	\centering
	\includegraphics[width=\linewidth]{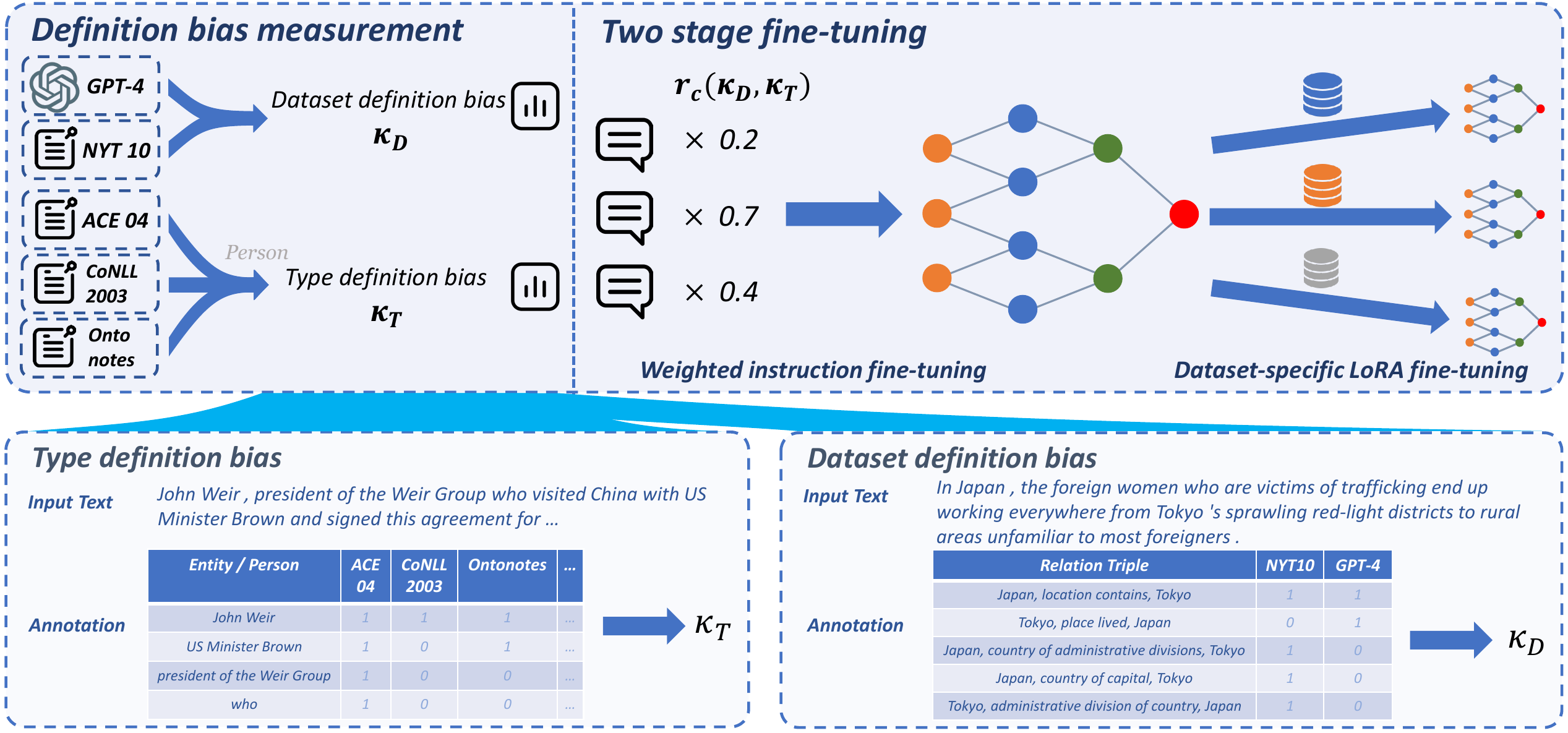}
	\caption{Our two-stage framework for alleviate definition bias. \textbf{Left:} we measure two kinds of definition bias with Fleiss' Kappa; \textbf{Right:} we first full-parameter fine-tune LLMs with measurement and then fine-tune with LoRA on specific dataset.}
	\label{fig:framework}
\end{figure*}

\renewcommand{\arraystretch}{1.4}
\begin{table}[!t]
    \small
    \centering
    \resizebox{\columnwidth}{!}{
    \begin{tabular}{lcccccc}
    \toprule
          \textbf{Model} & \multicolumn{3}{c}{\textbf{Llama-13b}}  & \multicolumn{3}{c}{\textbf{Flan-T5}}  \\
    \cmidrule(r){2-4}\cmidrule(r){5-7}
          \textbf{Source} & \textit{True} & \textit{Nickname} & \textit{Fake} & \textit{True} & \textit{Nickname} & \textit{Fake} \\
         \midrule
         \multicolumn{7}{c}{\cellcolor{lightgray!30}\textbf{\textit{Named Entity Recognition}}} \\
         
        \midrule
        \textbf{ACE 04} & 84.93 & 84.89 & 60.85 & 77.82 & 78.41 & 45.79\\ 
        \textbf{ACE 05} & 84.85 & 85.16 & 61.56 & 79.20 & 79.59 & 44.10\\
        \textbf{CoNLL 03} & 81.02 & 80.87 & 73.34 & 78.94 & 78.84 & 69.23\\
        \textbf{Ontonotes} & 91.85 & 91.81 & 81.79 & 91.03 & 91.04 & 78.71\\
        \textbf{WikiANN en} & 89.54 & 89.65 & 81.43 & 76.26 & 76.07 & 66.08\\
        \textbf{TweetNER 7} & 68.92 & 69.11 & 66.19 & 68.35 & 68.45 & 60.44\\
        \textbf{WikiNeural} & 96.03 & 95.93 & 83.51 & 94.03 & 94.03 & 74.30\\
        \textbf{PolyglotNER} & 80.21 & 80.41 & 68.34 & 74.00 & 74.03 & 54.24\\
        \midrule
        avg & - & - &  -12.6 & - & - & -18.4\\
         \midrule
         
         \multicolumn{7}{c}{\cellcolor{lightgray!30}\textbf{\textit{Relation Extraction}}} \\
         \midrule

         \textbf{CoNLL 04} & 69.88 & 69.51 & 61.73 & 67.09 & 67.00 & 57.34 \\

         \textbf{NYT10} & 97.80 & 97.78 & 94.82 & 96.20 & 96.20 & 90.54 \\

         \textbf{NYT11}& 76.14 & 76.24 & 72.82 & 76.14 & 76.41 & 71.94 \\

        \textbf{GIDs} & 80.49 & 80.15 & 78.69 & 76.41 & 76.34 & 74.26 \\ 

         \textbf{WikiKBP} & 64.68 & 65.67 & 63.50 & 63.78 & 63.94 & 59.64 \\
         \midrule
        avg & - & - & -3.5 & - & - & -5.2\\
    \bottomrule
    \end{tabular}}
    \caption{Different extraction results obtained by prompting the source prompt tuning UIE with \textit{true}, \textit{nickname} and \textit{fake} source name.}
    \label{tab:source_prompt}
\end{table}

\renewcommand{\arraystretch}{1.3}
\begin{table}[!t]
    \centering
    \resizebox{\columnwidth}{!}{
    \begin{tabular}{lccc}
    \toprule
        \textbf{Dataset} & \textbf{Llama-chat-70B} & \textbf{GPT-3.5-Turbo} & \textbf{GPT-4} \\
    \midrule
        \textbf{ACE04} & 8.56 | 30.42 & 19.68 | 32.81 & 13.70 | \textbf{35.16} \\
        \textbf{ACE05} & 17.64 | 33.48 & 20.83 | 34.32 & 16.13 | \textbf{45.30}\\
        \textbf{CoNLL 03} & 33.89 | 49.36 & 39.70 | 55.90 & 46.66 | \textbf{64.99}\\
        \textbf{Ontonotes} & 11.86 | 27.56 & 22.14 | 28.83 & 31.70 | \textbf{40.57}\\
        \textbf{WikiANN en} & 32.87 | 50.00 & 50.83 | 57.90 & 51.57 | \textbf{59.03}\\
        \textbf{TweetNER 7} & 31.77 | 35.68 & 32.98 | 38.13 & 36.62 | \textbf{47.88}\\
        \textbf{WikiNeural} & 42.98 | 57.03 & 50.00 | 59.83 & 65.23 | \textbf{70.66}\\
        \textbf{PolyglotNER} & 21.44 | 30.91 & 42.20 | 44.88 & \textbf{45.14} | 43.23\\
    \midrule
        \textbf{CoNLL 04} & 3.36 | 18.77 & 9.22 | 23.86 & 24.62 | \textbf{29.86}\\
        \textbf{NYT10} & 2.97 | 13.17 & 2.13 | 13.64 & 16.67 | \textbf{20.13}\\
        \textbf{NYT11} & 2.03 | 5.33 & 1.93 | 6.50 & 8.00 | \textbf{12.00}\\
        \textbf{GIDs} & 11.36 | 7.92 & 7.89 | 19.45 & 6.82 | \textbf{24.54}\\ 
        \textbf{WikiKBP} & 18.55 | 29.56 & 17.25 | 32.41 & 25.00 | \textbf{45.85}\\
    \bottomrule
    \end{tabular}
    }
    \caption{Performance of Open-source LLM and close-source LLM on various information extraction task in (\textbf{zero-shot} | \textbf{few-shot}) settings.}
    \label{tab:LLM_fs_zs}
\end{table}
\subsection{Can UIE address definition bias?}
Following the source prompt setting described in Section~\ref{SourcePromptSettings}, we tuned Llama-13b and Flan-T5 with source prompt instructions and prompted them with three source settings.

Table~\ref{tab:source_prompt} displays the extraction results evaluated by F1 scores. Replacing the true source names with fake ones results in a drop in F1 scores across all NER and RE tasks, with an average decrease of 12.6/3.5 and 18.4/5.2, respectively. However, when true source names are replaced with nicknames, the results show virtually no difference. This significant performance gap highlights that UIE is unable to mitigate definition bias during the multi-task learning process. The implicit definition bias permeates the model, leading to inconsistent extraction results, even when the same extraction task instructions are given.

\subsection{Can LLM address definition bias?}
The performance of various models on different tasks is presented in Table~\ref{tab:LLM_fs_zs}. Among the evaluated models, GPT-4 stands out by achieving the best performance across almost all datasets in both zero-shot and few-shot settings. Furthermore, the few-shot settings, which incorporate similar cases from the same dataset into the context, enhance performance by an average of 9.82 compared to the zero-shot settings. This improvement underscores the capacity of in-context learning to partially mitigate definition bias.

Despite these advances, it remains challenging for conventional, off-the-shelf methods to reach the performance levels of fully supervised approaches. This discrepancy underscores the presence of a significant definition bias between datasets used for information extraction and those used for instruction fine-tuning. Additionally, applying LLMs to information extraction faces two primary limitations. First, the constraint on context length prevents the inclusion of all annotated cases within the context. Second, the definition bias across different information extraction datasets complicates the creation of comprehensive prompts that accurately describe the extraction tasks.

\section{Alleviate Definition Bias}\label{sec:framework}
In this section, we explore methods to enhance the information extraction capabilities of LLMs.

Based on the probing experiments and conclusions outlined in Sections~\ref{sec:probing_task} and \ref{sec:probing_exp}, we find that definition bias across different datasets significantly iooompacts the performance of UIE and LLMs. This indicates that a framework relying solely on a one-stage, parameter-free update is inadequate for addressing definition bias. To tackle this challenge, we introduce a two-stage fine-tuning framework. Moreover, by identifying and explicitly quantifying the two types of definition biases we have discovered, we can integrate these measurements into our fine-tuning framework, effectively reducing the influence of definition bias.

\subsection{Definition bias measurement}
First, we introduce \textbf{Fleiss' Kappa}, a statistical measure used to assess the reliability of agreement among multiple raters when they assign categorical ratings to a set of items. This tool is valuable in identifying and mitigating definition bias.
\begin{equation}
    \kappa = 1-\frac{1-p_0}{1-p_e}=\frac{p_0-p_e}{1-p_e}
\end{equation}

where $p_o$ denotes the \textit{Observed Agreement}, the proportion of times that the raters actually agree, and $p_e$ denotes the \textit{Expected Agreement}, which represents the agreement that could be expected purely by chance. 
Suppose there are $N$ cases for a task, and each data is labeled $n$ times, and $k$ is the number of categories. These can be calculated using the following formula. 
\begin{equation}
    p_e = \sum_{j=1}^kp_j^2,\ \ p_j=\frac{1}{Nn}\sum_{i=1}^Nn_{ij}
\end{equation}
\begin{equation}
    p_o = \frac1N\sum_{i=1}^Np_i,\ \ p_i=\frac{1}{n(n-1)}\sum_{j=1}^kn_{ij}(n_{ij} - 1)
\end{equation}
where $n_{ij}$ denotes the number of annotator that label case $i$ as category $j$.

Specifically, we focus on the definition bias in information extraction and divide the definition bias into two type: \textbf{\textit{dataset definition bias}} and \textbf{\textit{type definition bias}}. 

\paragraph{Dataset Definition Bias $\kappa_D$} recognized as the agreement between GPT4 and the annotation of dataset, serving as a measure of reliability for transforming information extraction into instruction tuning dataset. It is carried out by calculating Fleiss's Kappa between the GPT4 extraction results and the golden annotation of the dataset.

\paragraph{Type Definition Bias $\kappa_T$} considered as the agreement among information extraction datasets with the same type either entity or relationship, and serves as a metric to evaluate the reliability of these types in terms of consistent annotation.

\subsection{Bias-aware fine-tuning}\label{sec:baft}
Based on the probing experiments, inconsistencies in definitions across various datasets significantly impact the training process. However, the diversity among these datasets, including annotation types and text sources, helps to improve the performance of LLMs for IE tasks. Therefore, it is essential to adopt a precise method to assess the quality of different datasets to effectively guide the training process.

We fine-tune the LLMs with information extraction dataset through C-RLFT~\cite{wang2023openchat}, which enables leveraging mixed-quality training data. We define the quality of the training samples as metrics based on $\kappa_D$ and $\kappa_T$. Suppose there are $N$ entity or relation triples in a case, we calculate the coarse-grained rewards of each case $r_c(x_i,y_i)$ by the formula below. 

\begin{equation}
    r_c(x_i,y_i) = (1 + \kappa_D)\frac1{N}\sum^N_{i=1}\kappa_{T_i}
\end{equation}

\begin{figure*}[t]
	\centering
	\includegraphics[width=\linewidth]{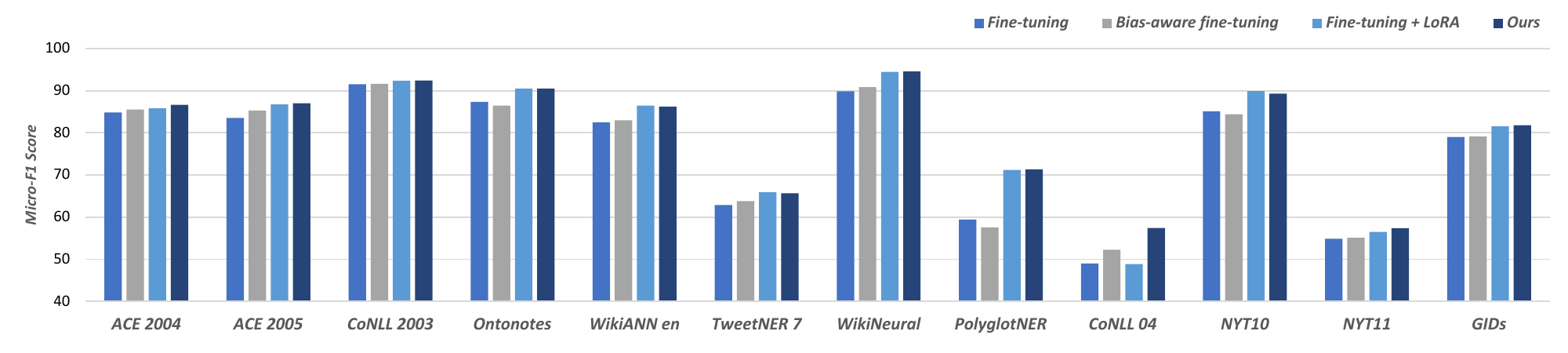}
	\caption{Ablation study on 12 information extraction dataset (NER and RE)}
	\label{fig:ablation_study}
\end{figure*}
\subsection{Task-specific bias mitigation}
To further enhance the performance of LLMs on ospecific information extraction datasets, we employ Low-Rank Adaptation (LoRA) for additional instruction tuning. We hypothesize that updates to the weights for a dataset possess a low intrinsic rank. This low intrinsic dimension adaptation can help mitigate the definition bias between a multi-task learning model and the dataset. Specifically, for a pre-trained weight matrix $W_0\in\mathbb{R}^{d\times k}$, we constrain its update with a low rank decomposition.
\begin{equation}
    h = W_0x + \Delta Wx = W_0x + BAx
\end{equation}
\begin{table}[!t]
    \centering
    \resizebox{\columnwidth}{!}{
    \begin{tabular}{lccc|c}
    \toprule
        \textbf{Dataset} & \textbf{UIE} & \textbf{USM} & \textbf{InstructUIE} & \textbf{Ours} \\
    \midrule
        \textbf{ACE 04} & 86.89 & \textbf{87.62} & - & 86.68 \\
        \textbf{ACE 05} & 85.78 & \textbf{87.14} & 86.66 & 87.05 \\
        \textbf{CoNLL 03} & 92.99 & \textbf{93.16} & 92.94 & 92.47 \\
        \textbf{Ontonotes} & - & - & 90.19 & \textbf{90.52} \\
        \textbf{WikiANN en} & - & - & 85.13 & \textbf{86.24} \\
        \textbf{TweetNER 7} & - & - & 64.97 & \textbf{65.70} \\
        \textbf{WikiNeural} & - & - & 91.36 & \textbf{94.59} \\
        \textbf{PolyglotNER} & - & - & 70.15 & \textbf{71.34} \\
    \midrule
        \textbf{CoNLL 04} & 75.00 & \textbf{78.84} & 78.48 & 57.46 \\
        \textbf{NYT10} & - & - & \textbf{90.47} & 89.35 \\
        \textbf{NYT11} & - & - & 56.06 & \textbf{57.38} \\
        \textbf{GIDs} & - & - & \textbf{81.98} & 81.83 \\
    \bottomrule
    \end{tabular}}
    \caption{Main result for comparing with other models on NER and RE tasks.}
    \label{tab:main_result}
\end{table}
where $\Delta Wx$ denotes the updatable parameters of $W_0$, and it can be constrained with a low-rank decomposition $\Delta W=BA$, where $B\in \mathbb{R}^{d\times r},A\in \mathbb{R}^{r\times k}$, and the rank $r \ll min(d,r)$. $W_0$ is frozen and does not receive gradient updates while $A$ and $B$ contain trainable parameters during training. 

In this stage, the model updates its parameter through further fine-tuning on a specific dataset to align the annotation.

\section{Experiments of Two-stage Framework}\label{sec:exp}

This section conducts experiments to validate the effectiveness of our two-stage fine-tuning framework. We select 11B Flan-T5~\cite{chung2022scaling} as our backbone model. The details of the experimental setup and comparison methods are described in the following parts.

\subsection{Experimental setup}
In bias-aware fine-tuning, we apply a sampling strategy to balance the dataset. In specific, we sample 10,000 cases from each dataset for training. In task-specific bias mitigation, we apply all examples for training. Further details can be found in Appendix~\ref{apd:training_settings}.

Our baseline models contain: UIE~\cite{lu-etal-2022-unified}, USM~\cite{lou2023universal}, and InstructUIE~\cite{wang2023instructuie}.

\subsection{Results}
Table~\ref{tab:main_result} presents the result on different dataset with baselines. Although our framework was trained on several information extraction datasets in the general domain, which might be considered unfair for comparing with baselines trained on other datasets, it achieves state-of-the-art on many datasets. It is worth noting that in datasets focusing exclusively on \texttt{person}, \texttt{location}, \texttt{organization} (as listed in Table~\ref{tab:label_list}), our framework achieves the best performance on WikiANN en, WikiNeural and PolyglotNER. This demonstrates the effectiveness of our framework in mitigating definition bias across different datasets.

\subsection{Experiment with two-stage fine-tuning}
To better improve the effectiveness of our two-stage fine-tuning framework, we conduct ablation study comparing with the following baseline: 
\begin{inparaenum}[1.]
    \item \textit{\textbf{Fine-tuning}}: fine-tuning the model with information extraction; 
    \item \textit{\textbf{Bias-aware fine-tuning}}: first stage fine-tuning in Section~\ref{sec:baft}; 
    \item \textit{\textbf{Fine-tuning+LoRA}}: data-specific instruction-tuning with LoRA on the weight of baseline 1; 
    \item \textit{\textbf{Ours}}: our two-stage fine-tuning framework.
\end{inparaenum}
The results are shown in Figure~\ref{fig:ablation_study}. In general, our framework nearly achieves the best performance compared to the baseline, demonstrating its effectiveness. By comparing baseline 1 and 2, it is proven that our bias-aware fine-tuning can alleviate definition bias among IE datasets and help models better align with GPT-4. It is also notable that two-stage fine-tuning consistently improves performance on specific datasets, attributed to the task-specific bias mitigation.

\section{Related Work}
\subsection{LLMs for information extraction}
Large language models have shown remarkable performance in instruction following \cite{openai2023gpt4}. To better align the natural instruction task from pre-trained and instruction tuning task, \citet{wei2023zero,wadhwa2023revisiting,zhang2023aligning} convert the structural information extraction task into natural instruction task such as question answering, multi-choice, etc.
While \citet{li2023codeie, guo2023retrieval} recast the structured output in the form of code to better leverage the LLMs of code to address the complex structure. Although LLMs show impressive performance in various information extraction tasks by designing fine-grained instruction, they still fail to address definition bias without further tuning.


\subsection{Universal information extraction}
Unified Information extraction, proposed by \citet{lu-etal-2022-unified}, uniformly encodes various information extraction tasks with a predefined structured extraction language (SEL), and enhances the common IE abilities via a large-scale pre-trained generation model. \citet{lou2023universal} further introduce USM to model different IE tasks, while \citet{wang2023instructuie} unified tasks into natural language instruction. GoLLIE converts the IE schema into code-style structural description and adds guidelines to improve zero-shot results \cite{sainz2023gollie}. However, they mainly focus on how to encode different extraction task into a uniform structure but fail to notice and detect the definition bias among various datasets.

\section{Conclusion}
In the paper, we propose the definition bias problem in information extraction task. We conduct several probing experiments to comprehensively demonstrate that existing methods cannot address definition bias. We then propose a multi-stage tuning framework, which consists of bias-aware fine-tuning and task-specific bias mitigation, to alleviate the definition bias in a specific dataset. Experimental results show that our framework is efficient in mitigating definition bias.



\section*{Limitation}
We systematically investigate definition bias in IE with devising a series of probing experiments. And we propose a multi-stage framework to mitigate definition bias in IE. However, there are still some limits of our probing experiment and the solution framework.

First, our probing experiment only focus on the definition bias among NER and RE tasks, which does not cover all the task in information extraction, which remains improvement for the future work.

Second, the performance of our solution framework is restricted by two main reason: 
\begin{inparaenum}[1)]
    \item more diverse dataset can be used for the bias-aware fine-tuning dataset; 
    \item the choice on backbone model also plays an important role in model performance. More experiments can more effectively validate the effectiveness of the proposed framework.
\end{inparaenum}

\section*{Ethic statement}
We hereby declare that all authors of this article are aware of and adhere to the provided ACL Code of Ethics and honor the code of conduct.

\paragraph{Use of Human Annotations}
Human annotations are only utilized in the early stages of methodological research to assess the feasibility of the proposed solution. All annotators have provided consent for the use of their data for research purposes. We guarantee the security of all annotators throughout the annotation process, and they are justly remunerated according to local standards. Human annotations are not employed during the evaluation of our method.

\paragraph{Risks}
The datasets used in the paper have been obtained from public sources and anonymized to protect against any offensive information. Though we have taken measures to do so, we cannot guarantee that the datasets do not contain any socially harmful or toxic language.

\bibliography{anthology,custom}

\begin{thebibliography}{38}
\expandafter\ifx\csname natexlab\endcsname\relax\def\natexlab#1{#1}\fi

\bibitem[{Al-Rfou et~al.(2015)Al-Rfou, Kulkarni, Perozzi, and Skiena}]{al2015polyglot}
Rami Al-Rfou, Vivek Kulkarni, Bryan Perozzi, and Steven Skiena. 2015.
\newblock Polyglot-ner: Massive multilingual named entity recognition.
\newblock In \emph{Proceedings of the 2015 SIAM International Conference on Data Mining}, pages 586--594. SIAM.

\bibitem[{Chung et~al.(2022)Chung, Hou, Longpre, Zoph, Tay, Fedus, Li, Wang, Dehghani, Brahma et~al.}]{chung2022scaling}
Hyung~Won Chung, Le~Hou, Shayne Longpre, Barret Zoph, Yi~Tay, William Fedus, Yunxuan Li, Xuezhi Wang, Mostafa Dehghani, Siddhartha Brahma, et~al. 2022.
\newblock Scaling instruction-finetuned language models.
\newblock \emph{arXiv preprint arXiv:2210.11416}.

\bibitem[{Dong et~al.(2023)Dong, Li, Dai, Zheng, Wu, Chang, Sun, Xu, Li, and Sui}]{dong2023survey}
Qingxiu Dong, Lei Li, Damai Dai, Ce~Zheng, Zhiyong Wu, Baobao Chang, Xu~Sun, Jingjing Xu, Lei Li, and Zhifang Sui. 2023.
\newblock \href {http://arxiv.org/abs/2301.00234} {A survey on in-context learning}.

\bibitem[{Ellis et~al.(2012)Ellis, Li, Griffitt, Strassel, and Wright}]{ellis2012linguistic}
Joe Ellis, Xuansong Li, Kira Griffitt, Stephanie~M Strassel, and Jonathan Wright. 2012.
\newblock Linguistic resources for 2013 knowledge base population evaluations.
\newblock In \emph{TAC}.

\bibitem[{Fleiss(1971)}]{fleiss1971measuring}
Joseph~L Fleiss. 1971.
\newblock Measuring nominal scale agreement among many raters.
\newblock \emph{Psychological bulletin}, 76(5):378.

\bibitem[{Guo et~al.(2023)Guo, Li, Jin, Liu, Zeng, Liu, Li, Yang, Bai, Guo et~al.}]{guo2023retrieval}
Yucan Guo, Zixuan Li, Xiaolong Jin, Yantao Liu, Yutao Zeng, Wenxuan Liu, Xiang Li, Pan Yang, Long Bai, Jiafeng Guo, et~al. 2023.
\newblock Retrieval-augmented code generation for universal information extraction.
\newblock \emph{arXiv preprint arXiv:2311.02962}.

\bibitem[{Hellström et~al.(2020)Hellström, Dignum, and Bensch}]{hellström2020bias}
Thomas Hellström, Virginia Dignum, and Suna Bensch. 2020.
\newblock \href {http://arxiv.org/abs/2004.00686} {Bias in machine learning -- what is it good for?}

\bibitem[{Hovy et~al.(2006)Hovy, Marcus, Palmer, Ramshaw, and Weischedel}]{hovy2006ontonotes}
Eduard Hovy, Mitch Marcus, Martha Palmer, Lance Ramshaw, and Ralph Weischedel. 2006.
\newblock Ontonotes: the 90\% solution.
\newblock In \emph{Proceedings of the human language technology conference of the NAACL, Companion Volume: Short Papers}, pages 57--60.

\bibitem[{Hu et~al.(2021)Hu, Shen, Wallis, Allen-Zhu, Li, Wang, Wang, and Chen}]{hu2021lora}
Edward~J. Hu, Yelong Shen, Phillip Wallis, Zeyuan Allen-Zhu, Yuanzhi Li, Shean Wang, Lu~Wang, and Weizhu Chen. 2021.
\newblock \href {http://arxiv.org/abs/2106.09685} {Lora: Low-rank adaptation of large language models}.

\bibitem[{Jat et~al.(2018)Jat, Khandelwal, and Talukdar}]{jat2018improving}
Sharmistha Jat, Siddhesh Khandelwal, and Partha Talukdar. 2018.
\newblock Improving distantly supervised relation extraction using word and entity based attention.
\newblock \emph{arXiv preprint arXiv:1804.06987}.

\bibitem[{Kojima et~al.(2022)Kojima, Gu, Reid, Matsuo, and Iwasawa}]{kojima2022large}
Takeshi Kojima, Shixiang~Shane Gu, Machel Reid, Yutaka Matsuo, and Yusuke Iwasawa. 2022.
\newblock Large language models are zero-shot reasoners.
\newblock \emph{Advances in neural information processing systems}, 35:22199--22213.

\bibitem[{Li et~al.(2022)Li, Tang, Nie, Wen, and Zhao}]{li2022learning}
Junyi Li, Tianyi Tang, Jian-Yun Nie, Ji-Rong Wen, and Wayne~Xin Zhao. 2022.
\newblock Learning to transfer prompts for text generation.
\newblock In \emph{Proceedings of the 2022 Conference of the North American Chapter of the Association for Computational Linguistics: Human Language Technologies}, pages 3506--3518.

\bibitem[{Li et~al.(2023)Li, Sun, Tang, Yan, Wu, Huang, and Qiu}]{li2023codeie}
Peng Li, Tianxiang Sun, Qiong Tang, Hang Yan, Yuanbin Wu, Xuanjing Huang, and Xipeng Qiu. 2023.
\newblock Codeie: Large code generation models are better few-shot information extractors.
\newblock \emph{arXiv preprint arXiv:2305.05711}.

\bibitem[{Lou et~al.(2023)Lou, Lu, Dai, Jia, Lin, Han, Sun, and Wu}]{lou2023universal}
Jie Lou, Yaojie Lu, Dai Dai, Wei Jia, Hongyu Lin, Xianpei Han, Le~Sun, and Hua Wu. 2023.
\newblock \href {http://arxiv.org/abs/2301.03282} {Universal information extraction as unified semantic matching}.

\bibitem[{Lu et~al.(2022)Lu, Liu, Dai, Xiao, Lin, Han, Sun, and Wu}]{lu-etal-2022-unified}
Yaojie Lu, Qing Liu, Dai Dai, Xinyan Xiao, Hongyu Lin, Xianpei Han, Le~Sun, and Hua Wu. 2022.
\newblock \href {https://doi.org/10.18653/v1/2022.acl-long.395} {Unified structure generation for universal information extraction}.
\newblock In \emph{Proceedings of the 60th Annual Meeting of the Association for Computational Linguistics (Volume 1: Long Papers)}, pages 5755--5772, Dublin, Ireland. Association for Computational Linguistics.

\bibitem[{Mitchell et~al.(2005)Mitchell, Strassel, Huang, and Zakhary}]{mitchell2005ace}
Alexis Mitchell, Stephanie Strassel, Shudong Huang, and Ramez Zakhary. 2005.
\newblock Ace 2004 multilingual training corpus.
\newblock \emph{Linguistic Data Consortium, Philadelphia}, 1:1--1.

\bibitem[{OpenAI(2022)}]{openai2022chatgpt}
OpenAI. 2022.
\newblock \href {https://openai.com/blog/chatgpt} {Chatgpt}.

\bibitem[{OpenAI(2023)}]{openai2023gpt4}
OpenAI. 2023.
\newblock \href {http://arxiv.org/abs/2303.08774} {Gpt-4 technical report}.

\bibitem[{Ouyang et~al.(2022)Ouyang, Wu, Jiang, Almeida, Wainwright, Mishkin, Zhang, Agarwal, Slama, Ray, Schulman, Hilton, Kelton, Miller, Simens, Askell, Welinder, Christiano, Leike, and Lowe}]{ouyang2022training}
Long Ouyang, Jeff Wu, Xu~Jiang, Diogo Almeida, Carroll~L. Wainwright, Pamela Mishkin, Chong Zhang, Sandhini Agarwal, Katarina Slama, Alex Ray, John Schulman, Jacob Hilton, Fraser Kelton, Luke Miller, Maddie Simens, Amanda Askell, Peter Welinder, Paul Christiano, Jan Leike, and Ryan Lowe. 2022.
\newblock \href {http://arxiv.org/abs/2203.02155} {Training language models to follow instructions with human feedback}.

\bibitem[{Pan et~al.(2017)Pan, Zhang, May, Nothman, Knight, and Ji}]{pan2017cross}
Xiaoman Pan, Boliang Zhang, Jonathan May, Joel Nothman, Kevin Knight, and Heng Ji. 2017.
\newblock Cross-lingual name tagging and linking for 282 languages.
\newblock In \emph{Proceedings of the 55th Annual Meeting of the Association for Computational Linguistics (Volume 1: Long Papers)}, pages 1946--1958.

\bibitem[{Riedel et~al.(2010)Riedel, Yao, and McCallum}]{riedel2010modeling}
Sebastian Riedel, Limin Yao, and Andrew McCallum. 2010.
\newblock Modeling relations and their mentions without labeled text.
\newblock In \emph{Machine Learning and Knowledge Discovery in Databases: European Conference, ECML PKDD 2010, Barcelona, Spain, September 20-24, 2010, Proceedings, Part III 21}, pages 148--163. Springer.

\bibitem[{Roth and Yih(2004)}]{roth2004linear}
Dan Roth and Wen-tau Yih. 2004.
\newblock A linear programming formulation for global inference in natural language tasks.
\newblock In \emph{Proceedings of the eighth conference on computational natural language learning (CoNLL-2004) at HLT-NAACL 2004}, pages 1--8.

\bibitem[{Sainz et~al.(2023)Sainz, García-Ferrero, Agerri, de~Lacalle, Rigau, and Agirre}]{sainz2023gollie}
Oscar Sainz, Iker García-Ferrero, Rodrigo Agerri, Oier~Lopez de~Lacalle, German Rigau, and Eneko Agirre. 2023.
\newblock \href {http://arxiv.org/abs/2310.03668} {Gollie: Annotation guidelines improve zero-shot information-extraction}.

\bibitem[{Sang and De~Meulder(2003)}]{sang2003introduction}
Erik~F Sang and Fien De~Meulder. 2003.
\newblock Introduction to the conll-2003 shared task: Language-independent named entity recognition.
\newblock \emph{arXiv preprint cs/0306050}.

\bibitem[{Song et~al.(2020)Song, Tan, Qin, Lu, and Liu}]{song2020mpnet}
Kaitao Song, Xu~Tan, Tao Qin, Jianfeng Lu, and Tie-Yan Liu. 2020.
\newblock Mpnet: Masked and permuted pre-training for language understanding.
\newblock \emph{arXiv preprint arXiv:2004.09297}.

\bibitem[{Su et~al.(2022)Su, Murtadha, Pan, Hou, Sun, Huang, Wen, and Liu}]{su2022global}
Jianlin Su, Ahmed Murtadha, Shengfeng Pan, Jing Hou, Jun Sun, Wanwei Huang, Bo~Wen, and Yunfeng Liu. 2022.
\newblock \href {http://arxiv.org/abs/2208.03054} {Global pointer: Novel efficient span-based approach for named entity recognition}.

\bibitem[{Takanobu et~al.(2019)Takanobu, Zhang, Liu, and Huang}]{takanobu2019hierarchical}
Ryuichi Takanobu, Tianyang Zhang, Jiexi Liu, and Minlie Huang. 2019.
\newblock A hierarchical framework for relation extraction with reinforcement learning.
\newblock In \emph{Proceedings of the AAAI conference on artificial intelligence}, volume~33, pages 7072--7079.

\bibitem[{Team et~al.(2023)Team, Anil, Borgeaud, Wu, Alayrac, Yu, Soricut, Schalkwyk, Dai, and Hauth}]{geminiteam2023gemini}
Gemini Team, Rohan Anil, Sebastian Borgeaud, Yonghui Wu, Jean-Baptiste Alayrac, Jiahui Yu, Radu Soricut, Johan Schalkwyk, Andrew~M. Dai, and Anja Hauth. 2023.
\newblock \href {http://arxiv.org/abs/2312.11805} {Gemini: A family of highly capable multimodal models}.

\bibitem[{Tedeschi et~al.(2021)Tedeschi, Maiorca, Campolungo, Cecconi, and Navigli}]{tedeschi-etal-2021-wikineural-combined}
Simone Tedeschi, Valentino Maiorca, Niccol{\`o} Campolungo, Francesco Cecconi, and Roberto Navigli. 2021.
\newblock \href {https://doi.org/10.18653/v1/2021.findings-emnlp.215} {{W}iki{NE}u{R}al: {C}ombined neural and knowledge-based silver data creation for multilingual {NER}}.
\newblock In \emph{Findings of the Association for Computational Linguistics: EMNLP 2021}, pages 2521--2533, Punta Cana, Dominican Republic. Association for Computational Linguistics.

\bibitem[{Touvron et~al.(2023)Touvron, Martin, Stone, Albert, Almahairi, Babaei, Bashlykov, Batra, Bhargava, Bhosale et~al.}]{touvron2023llama}
Hugo Touvron, Louis Martin, Kevin Stone, Peter Albert, Amjad Almahairi, Yasmine Babaei, Nikolay Bashlykov, Soumya Batra, Prajjwal Bhargava, Shruti Bhosale, et~al. 2023.
\newblock Llama 2: Open foundation and fine-tuned chat models.
\newblock \emph{arXiv preprint arXiv:2307.09288}.

\bibitem[{Ushio et~al.(2022)Ushio, Neves, Silva, Barbieri, and Camacho-Collados}]{ushio2022named}
Asahi Ushio, Leonardo Neves, Vitor Silva, Francesco Barbieri, and Jose Camacho-Collados. 2022.
\newblock Named entity recognition in twitter: A dataset and analysis on short-term temporal shifts.
\newblock \emph{arXiv preprint arXiv:2210.03797}.

\bibitem[{Wadhwa et~al.(2023)Wadhwa, Amir, and Wallace}]{wadhwa2023revisiting}
Somin Wadhwa, Silvio Amir, and Byron~C Wallace. 2023.
\newblock Revisiting relation extraction in the era of large language models.
\newblock \emph{arXiv preprint arXiv:2305.05003}.

\bibitem[{Walker et~al.(2006)Walker, Strassel, Medero, and Maeda}]{walker2006ace}
Christopher Walker, Stephanie Strassel, Julie Medero, and Kazuaki Maeda. 2006.
\newblock Ace 2005 multilingual training corpus.
\newblock \emph{Linguistic Data Consortium, Philadelphia}, 57:45.

\bibitem[{Wang et~al.(2023{\natexlab{a}})Wang, Cheng, Zhan, Li, Song, and Liu}]{wang2023openchat}
Guan Wang, Sijie Cheng, Xianyuan Zhan, Xiangang Li, Sen Song, and Yang Liu. 2023{\natexlab{a}}.
\newblock Openchat: Advancing open-source language models with mixed-quality data.
\newblock \emph{arXiv preprint arXiv:2309.11235}.

\bibitem[{Wang et~al.(2023{\natexlab{b}})Wang, Zhou, Zu, Xia, Chen, Zhang, Zheng, Ye, Zhang, Gui, Kang, Yang, Li, and Du}]{wang2023instructuie}
Xiao Wang, Weikang Zhou, Can Zu, Han Xia, Tianze Chen, Yuansen Zhang, Rui Zheng, Junjie Ye, Qi~Zhang, Tao Gui, Jihua Kang, Jingsheng Yang, Siyuan Li, and Chunsai Du. 2023{\natexlab{b}}.
\newblock \href {http://arxiv.org/abs/2304.08085} {Instructuie: Multi-task instruction tuning for unified information extraction}.

\bibitem[{Wei et~al.(2023)Wei, Cui, Cheng, Wang, Zhang, Huang, Xie, Xu, Chen, Zhang et~al.}]{wei2023zero}
Xiang Wei, Xingyu Cui, Ning Cheng, Xiaobin Wang, Xin Zhang, Shen Huang, Pengjun Xie, Jinan Xu, Yufeng Chen, Meishan Zhang, et~al. 2023.
\newblock Zero-shot information extraction via chatting with chatgpt.
\newblock \emph{arXiv preprint arXiv:2302.10205}.

\bibitem[{Xie et~al.(2021)Xie, Liang, Liu, Huang, Huang, and Xiao}]{xie-etal-2021-revisiting}
Chenhao Xie, Jiaqing Liang, Jingping Liu, Chengsong Huang, Wenhao Huang, and Yanghua Xiao. 2021.
\newblock \href {https://doi.org/10.18653/v1/2021.acl-long.277} {Revisiting the negative data of distantly supervised relation extraction}.
\newblock In \emph{Proceedings of the 59th Annual Meeting of the Association for Computational Linguistics and the 11th International Joint Conference on Natural Language Processing (Volume 1: Long Papers)}, pages 3572--3581, Online. Association for Computational Linguistics.

\bibitem[{Zhang et~al.(2023)Zhang, Guti{\'e}rrez, and Su}]{zhang2023aligning}
Kai Zhang, Bernal~Jim{\'e}nez Guti{\'e}rrez, and Yu~Su. 2023.
\newblock Aligning instruction tasks unlocks large language models as zero-shot relation extractors.
\newblock \emph{arXiv preprint arXiv:2305.11159}.

\end{thebibliography}
\bibliographystyle{acl_natbib}
\newpage
\appendix

\section{Further Details about Probing Experiments}
In this section, we further introduce the details on three probing experiments, including experiment settings and additional experiment results.

\subsection{Fully supervised settings}\label{apd:fully_supervised}
In fully supervised settings, we trained GlobalPointer~\cite{su2022global} and \textsc{ReRe}~\cite{xie-etal-2021-revisiting} for each task in 20 epoches, with a learning rate of 2e-5 and a batch size of 32.

The filter result with a similar semantic filtering process is shown in Table~\ref{tab:data_filter_ner} and Table~\ref{tab:data_filter_re}. While the cross-validation result without the semantic similarity filter process is shown in Table~\ref{tab:ner_wo_filter} and Table~\ref{tab:re_wo_filter}.

\renewcommand{\arraystretch}{1.3}
\begin{table}[t]
    \centering
    \resizebox{\columnwidth}{!}{
    \begin{threeparttable}
    \begin{tabular}{ccccccccc}
    \toprule
          & \darkercell \textbf{A04}\tnote{1} & \lightercell \textbf{A05}\tnote{2} & \darkercell \textbf{C03}\tnote{3} & \lightercell \textbf{Ont}\tnote{4} & \darkercell \textbf{Wie}\tnote{5} & \lightercell \textbf{TN7}\tnote{6} & \darkercell \textbf{WiN}\tnote{7} & \lightercell \textbf{PoN}\tnote{8} \\
          \darkercell \textbf{A04}\tnote{1} & & 451 & 1299 & 3404 & 1886 & 121 & 2104 & 1260\\
          
         \lightercell \textbf{A05}\tnote{2} & 489 & & 1112 & 4649 & 3552 & 204 & 1954 & 1752\\
         
         \darkercell \textbf{C03}\tnote{3} &  71 & 67 & &  409  & 2576 & 22 & 1128 & 550\\
         
         \lightercell \textbf{Ont}\tnote{4} & 569 & 770 & 1149   &  & 4194 & 180 & 2880 & 2504 \\ 
         
         \darkercell \textbf{Wie}\tnote{5} & 248 & 310 & 2338 & 2086 &  & 279 & 7752 & 6957\\
         
         \lightercell \textbf{TN7}\tnote{6} & 222 & 328 & 1486 & 1736 & 3016 &  & 2397 & 1648\\

         \darkercell \textbf{WiN}\tnote{7} & 387 & 407 & 2239 & 2843 & 8230 & 229 &  & 7624\\

         \lightercell \textbf{PoN}\tnote{8} & 558 & 669 & 2795 & 4918 & 9597 & 387 & 10869 & \\
         \midrule
         \textbf{\#Tot} & 812 & 1060 & 3453 & 8262 & 10000 & 576 & 11597 & 10000\\
         \bottomrule
    \end{tabular}

    \begin{tablenotes}
    \setlength{\multicolsep}{0cm}
    \begin{multicols}{3}
        \item[1] ACE 2004 
        \item[2] ACE 2005 
        \item[3] CoNLL 2003
        \item[4] Ontonotes 
        \item[5] WikiANN en 
        \item[6] TweetNER 7
        \item[7] WikiNeural 
        \item[8] PolyglotNER
    \end{multicols}
    \end{tablenotes}
    \end{threeparttable}}
    \caption{Filter result in NER task.}
    \label{tab:data_filter_ner}
\end{table}

\renewcommand{\arraystretch}{1.55}
\begin{table}[t]
    \resizebox{\columnwidth}{!}{
    \begin{tabular}{cccccc}
        \toprule
          & \darkercell \textbf{CoNLL 04} & \lightercell \textbf{NYT10} & \darkercell \textbf{NYT11} & \lightercell \textbf{GIDs} & \darkercell \textbf{WikiKBP} \\
          \darkercell \textbf{CoNLL 04} & & 1935 & 150 & 1879 & 72 \\
          
         \lightercell \textbf{NYT10} & 242 & & 344 & 3977 & 160 \\
         
         \darkercell \textbf{NYT11} & 189 & 4032 & & 3608 & 150 \\
         
         \lightercell \textbf{GIDs} & 52 & 887 & 83 & & 79 \\
         
         \darkercell \textbf{WikiKBP} & 3 & 71 & 8 & 70 &  \\

         \midrule 
         \textbf{\#Tot} & 288 & 5000 & 369 & 4307 & 182\\
         \bottomrule
    \end{tabular}
    }
    \caption{Filter result in NER task.}
    \label{tab:data_filter_re}
\end{table}

\renewcommand{\arraystretch}{1.5}
\begin{table}[t]
    \centering
    \resizebox{\columnwidth}{!}{
    \begin{threeparttable}
    \begin{tabular}{ccccccccc}
    \toprule
          & \darkercell \textbf{A04}\tnote{1} & \lightercell \textbf{A05}\tnote{2} & \darkercell \textbf{C03}\tnote{3} & \lightercell \textbf{Ont}\tnote{4} & \darkercell \textbf{Wie}\tnote{5} & \lightercell \textbf{TN7}\tnote{6} & \darkercell \textbf{WiN}\tnote{7} & \lightercell \textbf{PoN}\tnote{8} \\
          \darkercell \textbf{A04}\tnote{1} & \cellcolor{purple!\fpeval{8510/85.10}}85.10 & \cellcolor{purple!\fpeval{8154/84.45}}81.54 & \cellcolor{purple!\fpeval{3936/93.62}}39.36 & \cellcolor{purple!\fpeval{3115/93.21}}31.15 & \cellcolor{purple!\fpeval{4170/86.60}}41.70 & \cellcolor{purple!\fpeval{3565/77.91}}35.65 & 
          \cellcolor{purple!\fpeval{2970/95.21}}29.70 &
          \cellcolor{purple!\fpeval{2110/77.77}}21.10\\
          
         \lightercell \textbf{A05}\tnote{2} & \cellcolor{purple!\fpeval{8287/85.10}}82.87 & \cellcolor{purple!\fpeval{8445/84.45}}84.45 & \cellcolor{purple!\fpeval{4031/93.62}}40.31 & \cellcolor{purple!\fpeval{3005/93.21}}30.05 & \cellcolor{purple!\fpeval{4025/86.60}}40.25 & \cellcolor{purple!\fpeval{3472/77.91}}34.72 &
         \cellcolor{purple!\fpeval{2918/95.21}}29.18 &
         \cellcolor{purple!\fpeval{2061/77.77}}20.61\\
         
         \darkercell \textbf{C03}\tnote{3} & \cellcolor{purple!\fpeval{2563/86.18}}25.63 & \cellcolor{purple!\fpeval{1953/85.63}}19.53 & \cellcolor{purple!\fpeval{9219/92.19}}92.19 & \cellcolor{purple!\fpeval{6111/92.26}}61.11 & \cellcolor{purple!\fpeval{5506/86.60}}55.06 & \cellcolor{purple!\fpeval{6826/77.91}}68.26 &
         \cellcolor{purple!\fpeval{8889/95.21}}88.89&
          \cellcolor{purple!\fpeval{5148/77.77}}51.48\\
         
         \lightercell \textbf{Ont}\tnote{4} & \cellcolor{purple!\fpeval{3504/85.29}}35.04 & \cellcolor{purple!\fpeval{2589/84.86}}25.89 & \cellcolor{purple!\fpeval{5949/93.62}}59.49 & \cellcolor{purple!\fpeval{8969/89.69}}89.69 & \cellcolor{purple!\fpeval{4516/86.60}}45.16 & \cellcolor{purple!\fpeval{3788/68.61}}37.88 &
         \cellcolor{purple!\fpeval{6220/95.21}}62.20&
          \cellcolor{purple!\fpeval{3742/77.77}}37.42\\
         
         \darkercell \textbf{Wie}\tnote{5} & \cellcolor{purple!\fpeval{1918/86.18}}19.18 & \cellcolor{purple!\fpeval{1411/85.63}}14.11 & \cellcolor{purple!\fpeval{6456/93.62}}64.56 & \cellcolor{purple!\fpeval{3922/92.26}}39.22 & \cellcolor{purple!\fpeval{8660/86.60}}86.60 & \cellcolor{purple!\fpeval{5952/77.91}}59.52 &
         \cellcolor{purple!\fpeval{6452/95.21}}64.52&
          \cellcolor{purple!\fpeval{3928/77.77}}39.28\\
         
         \lightercell \textbf{TN7}\tnote{6} & \cellcolor{purple!\fpeval{2478/87.87}}24.78 & \cellcolor{purple!\fpeval{1910/87.20}}19.10 & \cellcolor{purple!\fpeval{7161/94.65}}71.61 & \cellcolor{purple!\fpeval{4684/92.40}}46.84 & \cellcolor{purple!\fpeval{5862/89.83}}58.62 & \cellcolor{purple!\fpeval{6339/63.39}}63.39 &
         \cellcolor{purple!\fpeval{7417/95.21}}74.17 &
          \cellcolor{purple!\fpeval{4704/79.69}}47.04\\

         \darkercell \textbf{WiN}\tnote{7} &
         \cellcolor{purple!\fpeval{2410/86.18}}24.10 &
         \cellcolor{purple!\fpeval{1877/85.63}}18.77 &
         \cellcolor{purple!\fpeval{7876/93.62}}78.76 &
         \cellcolor{purple!\fpeval{5869/92.26}}58.69 &
         \cellcolor{purple!\fpeval{5541/86.60}}55.41 &
         \cellcolor{purple!\fpeval{6451/77.91}}64.51 &
         \cellcolor{purple!\fpeval{9521/95.21}}95.21 &
          \cellcolor{purple!\fpeval{5136/77.77}}51.36\\

         \lightercell \textbf{PoN}\tnote{8} &
         \cellcolor{purple!\fpeval{1449/86.18}}14.49 &
         \cellcolor{purple!\fpeval{1017/85.63}}10.17 &
         \cellcolor{purple!\fpeval{4630/95.21}}46.30 &
         \cellcolor{purple!\fpeval{3818/92.26}}38.18 &
         \cellcolor{purple!\fpeval{4026/86.60}}40.26 &
         \cellcolor{purple!\fpeval{3445/77.91}}34.45 &
         \cellcolor{purple!\fpeval{6945/95.21}}69.45 &
         \cellcolor{purple!\fpeval{7777/77.77}}77.77\\
         









         
         \bottomrule
    \end{tabular}

    \begin{tablenotes}
    \setlength{\multicolsep}{0cm}
    \begin{multicols}{4}
        \item[1] ACE 2004 
        \item[2] ACE 2005 
        \item[3] CoNLL 2003
        \item[4] Ontonotes 
        \item[5] WikiANN en 
        \item[6] TweetNER 7
        \item[7] WikiNeural 
        \item[8] PolyglotNER
    \end{multicols}
    \end{tablenotes}
    \end{threeparttable}}
    \caption{Definition bias between different NER tasks without similar semantic filtering.}
    \label{tab:ner_wo_filter}
\end{table}
\renewcommand{\arraystretch}{1.6}
\begin{table}[t]
    \footnotesize
    \centering
    \resizebox{\columnwidth}{!}{
    \begin{tabular}{cccccc}
        \toprule
          & \darkercell \textbf{CoNLL 04} & \lightercell \textbf{NYT10} & \darkercell \textbf{NYT11} & \lightercell \textbf{GIDs} & \darkercell \textbf{WikiKBP} \\
          \darkercell \textbf{CoNLL 04} & \cellcolor{purple!\fpeval{6112/61.12}}61.12 & \cellcolor{purple!\fpeval{917/89.09}}9.17 & \cellcolor{purple!\fpeval{754/66.16}}7.54 & - & \cellcolor{purple!24.37}24.37 \\
         \lightercell \textbf{NYT10} & \cellcolor{purple!\fpeval{1436/58.45}}14.36 & \cellcolor{purple!\fpeval{8968/89.68}}89.68 & \cellcolor{purple!\fpeval{5316/55.10}}53.16 & \cellcolor{purple!\fpeval{1517/62.21}}15.17 & \cellcolor{purple!\fpeval{3030/37.66}}30.30 \\
         \darkercell \textbf{NYT11} & \cellcolor{purple!\fpeval{878/59.28}}8.78 & \cellcolor{purple!\fpeval{8313/88.95}}83.13 & \cellcolor{purple!\fpeval{5682/56.82}}56.82 & \cellcolor{purple!\fpeval{1250/62.21}}12.50 & \cellcolor{purple!\fpeval{3140/38.64}}31.40 \\
         \lightercell \textbf{GIDs} & - & \cellcolor{purple!\fpeval{744/70.26}}7.44 & \cellcolor{purple!\fpeval{317/26.67}}3.17 & \cellcolor{purple!\fpeval{6512/65.12}}65.12 & \cellcolor{purple!\fpeval{4621/46.21}}51.67 \\
         \darkercell \textbf{WikiKBP} & \cellcolor{purple!0.00}0.00 & \cellcolor{purple!\fpeval{795/80.69}}7.95 & \cellcolor{purple!\fpeval{253/67.33}}2.53 & \cellcolor{purple!\fpeval{1878/62.21}}18.78 & \cellcolor{purple!\fpeval{3657/36.57}}36.57 \\
         
         \bottomrule
    \end{tabular}}
    \caption{Definition bias between different RE tasks without similar semantic filtering.}
    \label{tab:re_wo_filter}
\end{table}

\subsection{Source prompt settings}\label{apd:source_prompt}
To eliminate the instruction bias that different datasets focus on different types of entity or relation, we employ a task decomposition approach, which involves constructing separate instructions for every entity type or relationship type. It helps decompose a task instruction with many types into atomic task instruction, which can be shared across different datasets. Such a setting compels the UIE model to focus solely on the source name and apply the distinct extraction principle for different source prompts. Table~\ref{tab:task_decompose} show a case of task decomposition on NER.
\begin{table}[!h]
    \small
    \centering
    \begin{tabularx}{\columnwidth}{X}
        \toprule
        \multicolumn{1}{c}{\cellcolor{lightgray!30}\textbf{Original Extraction Instruction}} \\
        Instruction: Please list all entity words in the text that fit the category. Here's the category list: \\
        \textcolor{purple}{\underline{\textit{[\texttt{person},\texttt{organization},\texttt{location}]}}} \\
        And then output the result in the format of ```type1: entity1; type2: entity2; ...``` \\
        Input: \textcolor{darkblue}{\underline{\textit{[Input text for NER]}}} \\
        Output: \\
        \midrule
        \multicolumn{1}{c}{\cellcolor{lightgray!30}\textbf{Decomposed Extraction Instruction}} \\
        Instruction: Please list all entity words in the text that fit the category. Here's the category list: \\
        \textcolor{purple}{\underline{\textit{[\texttt{person}]}}} \\
        And then output the result in the format of ```type1: entity1; type2: entity2; ...``` \\
        Input: \textcolor{darkblue}{\underline{\textit{[Input text for NER]}}} \\
        Output: \\
        \\
        Instruction: Please list all entity words in the text that fit the category. Here's the category list: \\
        \textcolor{purple}{\underline{\textit{[\texttt{organization}]}}} \\
        And then output the result in the format of ```type1: entity1; type2: entity2; ...``` \\
        \textit{\textcolor{gray}{/*Input text*/}} \\
        Input: \textcolor{darkblue}{\underline{\textit{[Input text for NER]}}} \\
        Output: \\
        \\
        Instruction: Please list all entity words in the text that fit the category. Here's the category list: \\
        \textcolor{purple}{\underline{\textit{[\texttt{location}]}}} \\
        And then output the result in the format of ```type1: entity1; type2: entity2; ...``` \\
        Input: \textcolor{darkblue}{\underline{\textit{[Input text for NER]}}} \\
        Output: \\
        \bottomrule
    \end{tabularx}
    \caption{A case for decomposing NER tasks instruction which focus on the entity type: \texttt{person}, \texttt{organization} and \texttt{location}.}
    \label{tab:task_decompose}
\end{table}

\subsection{Large language model zero/few-shot settings}\label{apd:llm_fs_zs}
We use the prompt in Table~\ref{tab:prompt_template} for probing experiments and multi-stage fine-tuning, which consist of task description, output format, in-context learning cases and input text.
\begin{table}[!t]
    \small
    \centering
    \begin{tabularx}{\columnwidth}{X}
        \toprule
        \multicolumn{1}{c}{\cellcolor{lightgray!30}\textbf{Prompt for Named Entity Recognition}} \\
        \textit{\textcolor{gray}{/*Task prompt*/}} \\
        Instruction: Please list all entity words in the text that fit the category. Here's the category list: \\
        \textit{\textcolor{gray}{/*Entity type List*/}} \\
        \textcolor{purple}{\underline{\textit{[List of the entity type]}}} \\
        \textit{\textcolor{gray}{/*Output Format*/}} \\
        And then output the result in the format of ```type1: entity1; type2: entity2; ...``` \\
        \textit{\textcolor{gray}{/*In-context learning cases*/}} \\
        \\
        \textit{\textcolor{gray}{/*Input text*/}} \\
        Input: \textcolor{darkblue}{\underline{\textit{[Input text for NER]}}} \\
        Output: \\
        \midrule
        \multicolumn{1}{c}{\cellcolor{lightgray!30}\textbf{Prompt for Relation Extraction}} \\
        \textit{\textcolor{gray}{/*Task prompt*/}}\\
        Instruction: Given a sentence or paragraph, and a given relationship set that describe the relation between entities. Here's the relation set: \\
        \textit{\textcolor{gray}{/*Relation type List*/}} \\
        \textcolor{purple}{\underline{\textit{[List of the relationship type]}}} \\
        \textit{\textcolor{gray}{/*Output Format*/}} \\
        Output the result in the format of ```(subject1, relation1, object1), (subject2, relation2, object2), ...``` \\
        \textit{\textcolor{gray}{/*In-context learning cases*/}} \\
        \\
        \textit{\textcolor{gray}{/*Input text*/}} \\
        Input: \textcolor{darkblue}{\underline{\textit{[Input text for RE]}}} \\
        Output: \\
        \bottomrule
    \end{tabularx}
    \caption{Prompts for two type of information extraction task: NER and RE.}
    \label{tab:prompt_template}
\end{table}


\section{Further Details about Two-stage Instruction Fine-tuning}
\subsection{Fleiss' Kappa}\label{apd:fleiss_kappa}
Table~\ref{tab:fleiss_kappa_type} and \ref{tab:fleiss_kappa_dataset} show the $\kappa_T$ and $\kappa_D$ that measure the \textbf{\textit{type definition bias}} and \textbf{\textit{dataset definition bias}} in several IE datasets.
\renewcommand{\arraystretch}{1.3}
\begin{table}[!ht]
    \centering
    \begin{tabular}{p{0.5\columnwidth}c}
    \toprule
        \textbf{Type} & \textbf{Fleiss' Kappa} \\
    \midrule
         \multicolumn{2}{c}{\cellcolor{lightgray!30}\textit{Entity Type of Named Entity Recognition}} \\
         \textbf{person} & 0.414 \\
         \textbf{location} & 0.428 \\
         \textbf{organization} & 0.364 \\
         \textbf{facility} & 0.021 \\
    \midrule
    \multicolumn{2}{c}{\cellcolor{lightgray!30}\textit{Relation Type of Relation Extraction}} \\
         \textbf{place lived} & 0.473 \\
         \textbf{place of birth} & 0.467 \\
         \textbf{place of death} & 0.408 \\
         \textbf{children} & 0.333 \\
         \textbf{location contains} & 0.150 \\
         \textbf{person of company} & 0.359 \\
    \bottomrule
    \end{tabular}
    \caption{$\kappa_T$ measured with dataset annotation}
    \label{tab:fleiss_kappa_type}
\end{table}

\begin{table}[!ht]
    \centering
    \begin{tabular}{p{0.5\columnwidth}c}
    \toprule
        \textbf{Dataset} & \textbf{Fleiss' Kappa} \\
    \midrule
         \textbf{ACE 2004} & -0.648\\
         \textbf{ACE 2005} & -0.546\\
         \textbf{CoNLL 2003} & -0.350\\
         \textbf{Ontonotes} & -0.594\\
         \textbf{PolyglotNER} & -0.567\\
         \textbf{TweetNER7} & -0.521\\
         \textbf{WikiANN en} & -0.409\\
         \textbf{WikiNeural} & -0.293\\
    \midrule
         \textbf{conll04} & -0.701\\
         \textbf{GIDS} & -0.748\\
         \textbf{NYT10} & -0.799\\
         \textbf{NYT11} & -0.879\\
         \textbf{WikiKBP} & -0.541\\
    \bottomrule
    \end{tabular}
    \caption{$\kappa_D$ measured with dataset annotation and GPT-4}
    \label{tab:fleiss_kappa_dataset}
\end{table}

\subsection{Two-stage training settings}\label{apd:training_settings}
The training hyperparameters of our multi-stage framework are listed in Table~\ref{tab:hyper_parameters}.

\begin{table}
    \centering
    \begin{tabular}{cc}
    \toprule
         \textbf{Hyperparameters} & \textbf{Settings}\\
    \midrule
         \multicolumn{2}{c}{\cellcolor{lightgray!30}\textit{Bias-aware fine-tuning}} \\
         Learning rate & 1e-5 \\
         Epoch & 5 \\
         Batch size & 384\\
    \midrule
        \multicolumn{2}{c}{\cellcolor{lightgray!30}\textit{Dataset-specific Mitigation}} \\
         Learning rate & 1e-5 \\
         LoRA rank & 8 \\
         LoRA\_key & q,v\\
         Epoch & 10/30 \\
         Batch size & 256\\
    \bottomrule
    \end{tabular}
    \caption{Hyper-parameters for two-stage training with Flan-T5.}
    \label{tab:hyper_parameters}
\end{table}

\section{Detail Statistic on Training Datasets}
We use 13 datasets in named entity recognition and relation extraction. For NER task, the used dataset include ACE04~\cite{mitchell2005ace}, ACE05~\cite{walker2006ace}, CoNLL2003~\cite{sang2003introduction}, Ontonotes~\cite{hovy2006ontonotes}, PolyglotNER~\cite{al2015polyglot}, TweetNER~\cite{ushio2022named}, WikiNeural~\cite{tedeschi-etal-2021-wikineural-combined} and WikiANN~\cite{pan2017cross}. For RE task, the datasets used include CoNLL 2004~\cite{roth2004linear}, GIDS~\cite{jat2018improving}, NYT10~\cite{riedel2010modeling}, NYT11-HRL~\cite{takanobu2019hierarchical} and Wiki-KBP~\cite{ellis2012linguistic}.

The statistic of datasets are listed in Table~\ref{tab:dataset_number}. The pre-defined entity or relation types for each dataset are shown in Table~\ref{tab:label_list}.
\begin{table}[!htb]
    \centering
    \begin{tabular}{lccc}
        \toprule
        \textbf{Dataset} & \textbf{\#Train} & \textbf{\#Valid} & \textbf{\#Test} \\
        \midrule
        \multicolumn{4}{c}{\cellcolor{lightgray!30}\textit{Named Entity Recognition}}  \\
        \textbf{ACE 2004} & 6202 & 745 & 812 \\
        \textbf{ACE 2005} & 7299 & 971 & 1060 \\
        \textbf{CoNLL 03} & 14041 & 3250 & 3453 \\
        \textbf{Ontonotes} & 59924 & 8528 & 8262 \\
        \textbf{PolyglotNER} & 393982 & 10000 & 10000 \\
        \textbf{TweetNER 7} & 7111 & 886 & 576 \\
        \textbf{WikiANN en} & 20000 & 10000 & 10000 \\
        \textbf{WikiNeural} & 92720 & 11590 & 11597 \\
        \midrule
        
        \multicolumn{4}{c}{\cellcolor{lightgray!30}\textit{Relation Extraction}} \\
        \textbf{CoNLL 04} & 922 & 231 & 288 \\
        \textbf{GIDs} & 8526 & 1417 & 4307 \\
        \textbf{NYT10} & 56196 & 5000 & 5000 \\
        \textbf{NYT11} & 62648 & 149 & 369 \\
        \textbf{WikiKBP} & 79934 & 20 & 182\\

        \bottomrule
    \end{tabular}
    \caption{Detailed datasets statistic.}
    \label{tab:dataset_number}
\end{table}
\begin{table*}[!htb]
    \centering
    \begin{tabularx}{\textwidth}{lX}
        \toprule
        \textbf{Dataset} & \textbf{Annotation type} \\
        \midrule
        \multicolumn{2}{c}{\cellcolor{lightgray!30}\textit{Named Entity Recognition}}  \\
        \textbf{ACE 2004} & geographical social political, organization, person, location, facility, vehicle, weapon \\
        \midrule
        \textbf{ACE 2005} & organization, person, geographical social political, vehicle, location, weapon, facility \\
        \midrule
        \textbf{CoNLL 03} & location, else, organization, person \\
        \midrule
        \textbf{Ontonotes} & date, organization, person, geographical social political, national religious political, facility, cardinal, location, work of art, law, event, product, ordinal, percent, time, quantity, money, language \\
        \midrule
        \textbf{PolyglotNER} & location, person, organization \\
        \midrule
        \textbf{TweetNER 7} & group, creative work, person, event, product, location, corporation \\
        \midrule
        \textbf{WikiANN en} & location, person, organization \\
        \midrule
        \textbf{WikiNeural} & location, person, organization \\
        \midrule
        
        \multicolumn{2}{c}{\cellcolor{lightgray!30}\textit{Relation Extraction}} \\
        \textbf{CoNLL 04} & company founded place, location contains, place lived, person of company, kill \\
        \midrule
        \textbf{GIDs} & place of death, place of birth, education degree, education institution \\
        \midrule
        \textbf{NYT10} & ethnicity, place lived, geographic distribution, company industry, country of administrative divisions, administrative division of country, location contains, person of company, profession, ethnicity of people, company shareholder among major shareholders, sports team of location, religion, neighborhood of, company major shareholders, place of death, nationality, children, company founders, company founded place, country of capital, company advisors, sports team location of teams, place of birth \\
        \midrule
        \textbf{NYT11} & nationality, country capital, place of death, children, location contains, place of birt, place lived, administrative division of country, country of administrative divisions, company, neighborhood of, company founders \\
        \midrule
        \textbf{WikiKBP} & parent, children, person of company, place of birth, place of death, place lived, religion \\

        \bottomrule
    \end{tabularx}
    \caption{The type of entity or relationship in each dataset.}
    \label{tab:label_list}
\end{table*}

\end{document}